\documentclass[conference]{IEEEtran}
\IEEEoverridecommandlockouts
\usepackage{cite}
\usepackage{amsmath,amssymb,amsfonts}
\usepackage{algorithm,algorithmic}
\usepackage{graphicx}
\usepackage{textcomp}
\usepackage{xcolor}
\def\BibTeX{{\rm B\kern-.05em{\sc i\kern-.025em b}\kern-.08em
    T\kern-.1667em\lower.7ex\hbox{E}\kern-.125emX}}
\begin{document}

\title{Sample-Rank: Weak Multi-Objective Recommendations Using Rejection Sampling\\
%{\footnotesize \textsuperscript{*}Note: Sub-titles are not captured in Xplore and
%should not be used}
%\thanks{Identify applicable funding agency here. If none, delete this.}
}

\author{\IEEEauthorblockN{Abhay Shukla\textsuperscript{*}\thanks{\textsuperscript{*}This research was carried out during their tenure at Swiggy.}}
	\IEEEauthorblockA{
		\textit{Swiggy}\\
		Bengaluru, India \\
		silpara@gmail.com}
	\and
	\IEEEauthorblockN{Jairaj Sathyanarayana}
	\IEEEauthorblockA{
		\textit{Swiggy}\\
		Bengaluru, India \\
		jairaj.s@swiggy.in}
	\and
	\IEEEauthorblockN{Dipyaman Banerjee\textsuperscript{*}}
	\IEEEauthorblockA{
		\textit{Swiggy}\\
		Bengaluru, India \\
		dipyaman@gmail.com}
}

\maketitle

\begin{abstract}
Online food ordering marketplaces are multi-stakeholder systems where recommendations impact the experience and growth of each participant in the system. A recommender system in this setting has to encapsulate the objectives and constraints of different stakeholders in order to find utility of an item for recommendation. Constrained-optimization based approaches to this problem typically involve complex formulations and have high computational complexity in production settings involving millions of entities. Simplifications and relaxation techniques (for example, scalarization) help but introduce sub-optimality and can be time-consuming due to the amount of tuning needed. In this paper, we introduce a method involving multi-goal sampling followed by ranking for user-relevance (Sample-Rank), to nudge recommendations towards multi-objective (MO) goals of the marketplace. The proposed method’s novelty is that it reduces the MO recommendation problem to sampling from a desired multi-goal distribution then using it to build a production-friendly learning-to-rank (LTR) model. In offline experiments we show that we are able to bias recommendations towards MO criteria with acceptable trade-offs in metrics like AUC and NDCG. We also show results from a large-scale online A/B experiment where this approach gave a statistically significant lift of 2.64\% in average revenue per order (RPO) (objective \#1) with no drop in conversion rate (CR) (objective \#2) while holding the average last-mile traversed flat (objective \#3), vs. the baseline ranking method. This method also significantly reduces time to model development and deployment in MO settings and allows for trivial extensions to more objectives and other types of LTR models. 
\end{abstract}


\section{Introduction}
Recommendation systems in online food-ordering marketplaces (Swiggy, Zomato, etc.) help users find relevant items from restaurants which are then delivered by delivery partners. A few factors distinguish these marketplaces from regular ecommerce marketplaces. Firstly, the ordered items need to be delivered in a relatively short period of time-- in as little as 30 minutes as opposed to days. Secondly, the demand follows the breakfast-lunch-dinner cadence and the restaurants need to plan for staffing and availability of items to cater to these peaks. Thirdly, since most delivery partners are gig workers, the marketplace needs to ensure incentives to manage optimal staffing levels. In such a multi-stakeholder setting, recommenders need to frequently adapt to changes and accommodate business constraints. For example, recommenders may need to ensure fairness of demand across restaurants offering different cuisines while ensuring the cost-of-delivery from restaurants doesn’t exceed business-acceptable levels and providing a pre-set minimum number of deliveries to logged-in delivery partners. In such a situation, a unilateral focus on a single stakeholder may come at the expense of other participants and be detrimental to long term prospects of the marketplace.

Most online food ordering marketplaces need to serve the interests of at least 4 types of stakeholders. Customers (C) form the demand side of the equation and typically expect high-quality recommendations and service. Restaurants (R) provide the supply and lend themselves and their dishes to be surfaced in recommendations on the app. Delivery Partners (D) are incentivized to fulfill the orders within promised service-level agreements (SLA) while maintaining high levels of safety and quality of service. The System (S) is the marketplace that enables interactions between all of the above stakeholders and tries to optimize their objectives, including its own.

A principled way of solving this MO problem is to pose recommendations as a constrained optimization (CO) problem where one of the objectives is typically considered primary and optimized while others are set as constraints \cite{b1}, \cite{b2}. However, as more constraints are added, CO approaches become computationally expensive for real-world applications. To alleviate this, approaches which do relaxation and/or approximations of the original formulation have been proposed. But, formulating appropriate objective functions and converting business requirements into mathematical constraints can be a complex, time-consuming exercise.  

Given the difficulty CO based approaches pose in production settings, a popular technique is to reduce the MO optimization (MOO) to scalarization, where a weighted sum of the objectives is optimized. However, these weights are often manually determined and Pareto optimality is not guaranteed \cite{b4}. In some other applications, MOO has been accomplished by applying certain post-processing to scores obtained from recommender systems \cite{b3}, \cite{b7}, \cite{b12}, \cite{b19}, \cite{b20}.

In \cite{b5} Momma, M. et al. have addressed production concerns by taking the CO method applied to the Gradient Boosted Tree (GBT). However, this still involves modifying the CO problem using an Augmented Lagrangian and making changes to the loss function of the LTR models, to accomplish the MOO.

We propose a solution that addresses practical challenges of real-world MOO recommenders and scales to millions of records, with commonly used LTR models. Our solution does not require the MOO problem to be formulated as a CO or a scalarization problem. Rather, we propose splitting the problem into 2 stages. First stage involves using rejection sampling to extract a sample that encapsulates the multiple objectives to be optimized. Second stage achieves relevance-ranking using an LTR model. We call this method, Sample-Rank. While optimality is not guaranteed, we show that Sample-Rank results in nudging the recommendations towards the desired objectives while still maintaining user-relevance. 

In offline experiments we are able to bias recommendations towards MO goals with acceptable trade-offs in metrics like AUC and NDCG. We also show results from a large-scale online A/B experiment where recommendations with the proposed approach gave a statistically significant lift of 2.64\% in average revenue per order (objective \#1) with no drop in conversion rate (objective \#2) while holding the average last-mile traversed (LM) flat (objective \#3), vs. the baseline ranking method.

The remainder of the paper is structured as follows. In Section II, we discuss related work. In Section III, we give the necessary background and notation used in the rest of the paper. We describe Sample-Rank and Generalized Sample-Rank algorithms in Sections IV. In section V we describe the experiments, followed by results in Section VI. We conclude with a summary and possible directions for future work in Section VII.

\section{Related Work}
In this section we will give an overview of recommender systems optimizing multiple objectives, multistakeholder recommender systems, sampling methods in machine learning and learning to rank methods popular in the domain of recommendations.
\subsection{Multi-Objective Recommender Systems}

Multi-objective recommendation has been an active area of research evolving in numerous directions. A few to note are, a) optimizing weighted sum of multiple objectives into a single objective function called scalarization, b) optimizing a primary objective while setting others as the constraints of the optimization problem, c) post-processing the results with available information to serve multiple objectives and d) combining multiple model each learning a specific objective. Agarwal, D et al. \cite{b1} note that optimizing for clicks alone could divert users to low revenue entertainment sites leading to loss of revenue. They follow scalarization and constrained optimization with downstream metrics such as time spent as the primary objective to optimize. They study the trade-off between CTR and time spent between different formulations in offline experiments. Jambor, T et al. \cite{b3} take a post-processing approach and distinguish between predicted rating of an item and utility of the item. They post-process the predicted score by recommendation algorithm to optimize for additional objectives like promoting long tail items by posing the problem as a linear optimization problem. Abdollahpouri, H. et al. \cite{b12} also address the same problem by iteratively post-processing recommendation scores with diversity scores of candidate items to add to the user recommendation list. Das, A. et al. \cite{b7} also take a post processing optimization approach where they adjust output of the recommender system according to item profitability. In \cite{b6} Nguyen, P. et al. combine multiple models tuned for each objective separately such that the new combined score is close to predictions from independent models. 

\subsection{Multi-stakeholder Recommender Systems for Online Marketplaces}
When a user is not the sole stakeholder of a recommendation system, objectives relevant to other stakeholders, like fairness, delivery partner earnings, restaurant earnings, platform profitability etc., become important factors to consider. Multi-stakeholder recommendation (MSR) has emerged as the framework for addressing such settings by employing multi-objective optimization techniques discussed in II-A. A survey of problem settings and techniques can be found in  Abdollahpouri, H. et al. \cite{b11}.  In \cite{b20}, Abdollahpouri, H., use two approaches for MSR, 1) optimizing a regularized objective for recommendation and 2) post-processing ranking scores to make it multi-stakeholder aware. Surer, O. et al. \cite{b19} also post-process utility score for (user, item) pairs learned by recommendation algorithm to make it multi-stakeholder aware and use lagrangian relaxation to address computational complexity.

\subsection{Sampling Methods in Machine Learning}
Simple and stratified random sampling are commonly used sampling techniques which have been discussed extensively along with other traditional techniques by Cochran \cite{b14}. Often, the main purpose of traditional sampling methods is to obtain a smaller dataset while maintaining the structure \cite{b13}. In machine learning, special sampling methods have been developed to improve the classification performance of models, most notably for class imbalance problems \cite{b15}, \cite{b16}, \cite{b17}. Cheng, M. et al. in \cite{b21} describe a novel sampling strategy to select the most informative examples for efficiently training pairwise ranking models. The method of rejection sampling was formalized by von Neumann \cite{b18} and has been used since in a wide range of applications. Under certain assumptions of boundedness, rejection sampling can be used to generate samples from any closed form density function. The method proposed in this paper is directly inspired by the idea of rejection sampling.

\subsection{Learning to rank}
Learning to rank (L2R) methods are ubiquitous in modern recommendation systems. Pointwise, pairwise and listwise, the three popular L2R methods are described in \cite{b8}. LambdaMART is one of the best performing algorithms winning the Yahoo! Learning to Rank Challenge \cite{b9}, \cite{b10}. However, the method we propose in this paper is independent of the ranking method, therefore for simplicity we will use a pointwise approach using Gradient Boosting Trees (GBT). 

\section{Background}
Let $D$ denote the dataset which is a collection of feature and label pairs ($x_{i}$, $y_{i}$), i.e. ($x_{i}$, $y_{i}$) $\in$ $D$, where $x_i \in X, y_{i} \in Y$, additionally $X \in \mathbb{R}^{N \times n}$ and $Y \in \{0, 1\}^N$, where N is the size of dataset. H = \{C, R, D, S\} is the set of all the stakeholders affected by recommendations to C using ranking function $r$. Let $O_{H}(r(X, Y))$ denote the joint stakeholder objective function. Goal of a multi-stakeholder ranking system is to find the ranking function $r^*$ such that,
\begin{equation}
r^* = \underset{r}{arg\;max}\;O_{H}(r(X, Y))
\end{equation}
%$$r^* = \underset{r}{arg\;max}\;O_{H}(r(X, Y))$$

Note that, this is often a constrained optimization problem to ensure minimum business guarantees to the stakeholders. Also, if ranking function $r$ is parameterized by parameter $w$ then, subject to constraints, the objective is to find $w^*$ such that,
\begin{equation}
w^* = \underset{w}{arg\;max}\;O_{H}(r(X, Y, w))
\end{equation}
%$$w^* = \underset{w}{arg\;max}\;O_{H}(r(X, Y, w))$$
Often, this problem is simplified to maximizing user objectives with minimum guarantees promised to other stakeholders. 
\begin{equation}
\underset{w}{max}\;O_{C}(r(X, Y, w))
\end{equation}
s.t. 
\begin{equation}
\pi_{R}(r(X, Y, w)) \leq v_R
\end{equation}
\begin{equation}
\pi_{D}(r(X, Y, w)) \leq v_D
\end{equation}
\begin{equation}
\pi_{S}(r(X, Y, w)) \leq v_S
\end{equation}
%$$\underset{w}{max}\;O_{C}(r(X, Y, w))$$
%s.t. 
%$$\pi_{R}(r(X, Y, w)) \leq v_R$$
%$$\pi_{D}(r(X, Y, w)) \leq v_D$$
%$$\pi_{S}(r(X, Y, w)) \leq v_S$$

Where $\pi_{R}, v_R, \pi_{D} ,v_D,  \pi_{S}, v_S$ define restaurant, delivery partner and system business guarantee constraints on ranking function $r$. As noted earlier, this is a principled approach but hard to implement in many real world recommender systems due to their scale and frequently changing objectives. This often leads to approximations and relaxation of constraints. In this paper, we will use sampling to encapsulate stakeholder objectives in the data used for training the ranking model for recommendations. For convenience, the notation used in the rest of this paper is produced in Table \ref{tab1}.

% A table with adjusted row and column spacings
% \setlength sets the horizontal (column) spacing
% \arraystretch sets the vertical (row) spacing
\begingroup
\setlength{\tabcolsep}{2pt} % Default value: 6pt
\renewcommand{\arraystretch}{1.05} % Default value: 1
\begin{table}[htbp]
	\caption{Table of Notations}
	\begin{center}
		\begin{tabular}{l l}
			\hline
			\textbf{\textit{Notation$^{\mathrm{*}}$}}&\textbf{\textit{Description}}\\
			\hline
			$D, D^g$& Original and sampled data\\
			$X, Y$&Features, labels, $x \in X, y \in Y$\\
			$f(.), M$&Density function, maximum value of density function\\
			$f_k(.), M_{k}$&Density function of k-th component, and its maximum value\\
			$\mu, \sigma, \Sigma$&Mean, standard deviation and covariance matrix\\
			$\Delta$&Function determing change in parameters\\
			$\delta$&Determines amount of change in  parameter\\
			$U$&Uniform Distribution\\
			$r, w$&Ranking function and its parameters\\
			$t^2$&Squared Mahalanobis distance\\
			\hline
			\multicolumn{2}{l}{$^{\mathrm{*}}$Notation related to goal distribution or its components use}\\
			\multicolumn{2}{l}{super/sub-scripts $g$ and $k$ respectively.}
		\end{tabular}
		\label{tab1}
	\end{center}
\end{table}
\endgroup

\section{Proposed Approach}
We assume that feature space $X$ can be approximated by multivariate normal distribution $f$ and estimate the parameters $\mu$ and $\Sigma$ matrix using maximum likelihood estimation (MLE). We then postulate a goal distribution $f^{g}$, with parameters $\mu^{g}$ and $\Sigma^{g}$, such that it is desirable for multi-stakeholder objectives. The goal distribution parameters are determined by $\Delta$ functions for each parameter.  We do not assume any specific form of $\Delta$ function in the algorithm. Common choices may take the form of additive change, percentage change or shrinkage. E.g.\\

\begin{align}
\Delta_{\mu}({\mu}, {\delta}) &= \mu + \delta,\;-\infty < \delta < \infty&(additive\; change)\\
\Delta_{\mu}(\mu, \delta) &= \mu + \delta\mu,\;-\infty < \delta < \infty&(percentage\; change)\\
\Delta_{\Sigma}(\Sigma, \delta) &= \delta\Sigma,\;0 < \delta < 1 &(shrinkage)
\end{align}

An additive change in a component of $\mu$ can be interpreted as the required change in the component needed to  bias in recommendations towards desirable stakeholder objective, affected by the component. For example if the objective is improve the quality of restaurants in the recommendations, we may change the parameter corresponding to rating of restaurants, say $\mu_{restaurant\_rating}$, to $\mu_{restaurant\_rating}$ + $\delta_{restaurant\_rating}$ to obtain the goal distribution. For different values of $\delta$, we obtain different ranking models with a trade-off between NDCG and top@k metrics quantifying the effect produced in ranking. An appropriate model can be accordingly chosen. Effects of change in $\Sigma$ can be similarly interpreted.

After choosing appropriate parameters, we use the goal distribution for sampling the dataset $D$ to obtain sampled dataset $D^g$. Sampling is done using a modification over rejection sampling technique. Specifically, we have dropped the density function $f$ from the denominator in the reject-accept step. Note that in many cases the boundedness condition over $f^{g}(x)/f(x)$ of rejection sampling may not hold for goal distributions of choice but making this change in rejection sampling helps us avoid over-sampling points far from goal distribution parameters. With dataset $D^g$, we train a ranking model optimizing for user relevance. The sampling step encapsulates the desired effect of multiple-objectives by using the goal distribution, while learning-to-rank optimizes for user relevance. Details of sampling are given in Algorithm 1. Note that we can generalize this algorithm to the case of Gaussian Mixture Models by first assigning a given data point to a component distribution, followed by the rejection sampling procedure similar to Algorithm 1. Details of Generalized Sample-Rank are given in Algorithm 2.

\subsection{Sample-Rank}
\begin{algorithm}[H]
	\caption{(Sample-Rank)}
	\begin{algorithmic}[1]
	\STATE	Initialize $D^g\;to\;empty\;set$
	\STATE	Fit probability density $f(x)$ on D and get $M$
	\STATE	Define parameters of goal distribution $f^g$  as
		\begin{align*}
			\mu^{g} &= \Delta_{\mu}(\mu, \delta_{\mu})\\
			\Sigma^{g} &= \Delta_{\Sigma}(\Sigma, \delta_{\Sigma})
		\end{align*}
		\FOR{$(x_i, y_i) \in D$}
		\IF{$U(0, 1) < f^{g}(x_{i})/M$}
		\STATE $D^g = D^g \cup (x_i, y_i)$
		\ELSE
		\STATE reject $(x_i, y_i)$\\
		\ENDIF\\
		\ENDFOR
		\STATE Train ranking model $r(X, Y, w)$ using $D^g$
		\label{algo1}
	\end{algorithmic}
\end{algorithm}

\subsection{Generalized Sample-Rank}
\begin{algorithm}[H]
	\caption{(Generalized Sample-Rank)}
	\begin{algorithmic}[1]
		\STATE	Initialize $D^g\;to\;empty\;set$
		\STATE	Fit a gaussian mixture model with $p$ components, and get the maximum value for each component $f_k(x)$ denoted by $M_k$, where $k \in K$ and $K = \{1, 2, 3, 4, ..., p\}$
		\STATE	Define parameters of goal distribution $f^{g}_{k}$  as
		\begin{align*}
		\mu^{g}_{k} &= \Delta_{\mu}(\mu_{k}, \delta_{\mu,k})\\
		\Sigma^{g}_{k} &= \Delta_{\Sigma}(\Sigma_{k}, \delta_{\Sigma,k})
		\end{align*}
		\FOR{$(x_i, y_i) \in D$}
		\STATE Compute the squared Mahalanobis distance $t^{2}_k$ for each gaussian component k as,
		\begin{align*}
		t^{2}_{k} = (x_{i} - \mu^{g}_{k})^{T}{\Sigma^{g}}^{-1}_{k} (x_i - \mu^{g}_{k})
		\end{align*}
		\STATE Select the component distribution $k^*$ such that,
		\begin{align*}
		k^* = \underset{k \in K}{arg\;min}\;t^{2}_{k}
		\end{align*}
		\IF{$U(0, 1) < f^{g}_{k^*}(x_{i})/M_{k^*}$}
		\STATE $D^g = D^g \cup (x_i, y_i)$
		\ELSE
		\STATE reject $(x_i, y_i)$\\
		\ENDIF\\
		\ENDFOR
		\STATE Train ranking model $r(X, Y, w)$ using $D^g$
		\label{algo2}
	\end{algorithmic}
\end{algorithm}

Although the approach proposed in this paper is general but for illustration and experiments the stakeholders we consider are Customers (C), Restaurants (R) and Delivery Partners (D). Therefore, the postulated distribution will only consider parameters corresponding to features relevant to objectives of these stakeholders to obtain goal distribution parameters. E.g. items relevant to customers and with lower LM has direct effect on 1) customer satisfaction, due to relevance, and 2) delivery partners, since lower LM’s could mean lower delivery times and hence more deliveries and earnings to the delivery partners. Therefore in this case only the  parameters corresponding to LM and customer experience need to be changed to obtain goal distribution. 

Empirically, we observe that sampling from goal distribution 'nudges' the recommendations towards the desired multi-objective goals when compared to the baseline model. Therefore, we call this paradigm weak multi-objective recommendations.

The notation used for Sample-Rank and Generalized Sample-Rank algorithms, is described in Table \ref{tab1}.

\section{Experiments}
In this section, we provide details of the baseline model, evaluation strategy, metrics and dataset used for training and evaluation of Sample-Rank.

\subsection{Evaluation Strategy}
Since Sample-Rank uses a goal distribution to sample from the original data for training models, evaluating performance on the sampled test set will be biased and likely optimistic. In order to get an unbiased estimate of the performance of the algorithm, we evaluate it on a test set held out from the original data. 
\subsection{Baseline}
Baseline model is a pointwise LTR model using a GBT classifier trained on the original data. Since this model learns from the original data distribution, it is expected to reinforce biases and patterns present in the historical interactions on the platform.
\subsection{Evaluation Metrics}
We use AUC and NDCG as primary evaluation metrics. We also look at the impact in recommendations upto top-k positions as increment relative to baseline performance (top@k). We define incremental performance as Sample-Rank performance - baseline performance. top@k metrics give a sense of the nudging Sample-Rank is able to provide and help in choosing appropriate parameters to use for goal distribution. This is critical in online food ordering marketplaces since a nudge too extreme can lead to poor customer experience due to detrimental trade-offs with user-relevance.
\subsection{Dataset}
We use 3 weeks of transaction history with over 40 features. This data consists of all items seen by the customer in a session with the item that was ordered representing the positive example and the rest of the items constitute negative examples. Customers are assigned into training and test sets in 70:30 ratio. Splitting at customer level ensures that a given customer’s transactions are present exclusively either in training or test set but not both. For goal sampling, instead of considering (customer, item) pairs in transacted sessions independently, we select or reject the complete session. This is important since a successful transaction occurs in context of all the choices presented at that time. If the transacted item is rejected by the goal sampling then we only have negative examples left in the session and they add no relevant distinguishing information for ranking.
\subsection{Experiment Settings}
Now, we describe the goal sampling parameters used to evaluate Sample-Rank in offline and online experiments. We assume that multivariate normal distribution with single component is representative of the feature space. Also note that all changes in $\mu$ are of additive nature and we do not change $\Sigma$ in the goal distribution.
\subsubsection{Offline Experiment Settings}
We used (restaurant rating, item rating) and (LM, RPO) as the parameters for goal sampling in offline experiments. These settings are described in Table \ref{tab2}.
\begin{table}[htbp]
	\caption{Offline Experiment Settings}
	\begin{center}
	\begin{tabular}{c|c|c}
		\hline
		\textbf{\textit{Experiment}}& \textbf{\textit{Sampler deviation}}&\textbf{\textit{Sampler deviation}}\\
		&\textbf{\textit{(item rating, restaurant rating)}}&\textbf{\textit{(LM, RPO)}}\\
		\hline
		Baseline&[+0, +0]&[-0, +0]\\
		Goal 1&[+0.1, +0.1]&[-0, +20]\\
		Goal 2&[+0.2, +0.2]&[-0.5, +0]\\
		Goal 3&[+0.3, +0.3]&[-0.5, +20]\\
		Goal 4&[+0.4, +0.4]&[-0.75, +20]\\
		Goal 5&[+0.5, +0.5]&[-0.75, +40]\\
		\hline
	\end{tabular}
	\label{tab2}
	\end{center}
\end{table}

\subsubsection{Online Experiment Settings}
For the online experiment, we chose RPO to +20 and LM set to -0 as the sampling parameters. Note that this setting is analogous to Goal 1: [-0, +20] for (LM, RPO) in offline experiments. Baseline model details are explained in section V-B. An A/B experiment with 50\%-50\% split for both variants was conducted for a period of 9 days in a large market. The results of the experiment are discussed in section VI.
\section{Results}
We present the results of different experiment settings discussed in Section V-E. For offline experiments, we measure the impact of using Sample-Rank on AUC, NDCG and top@k metrics. Online experiment measures whether we are able to achieve the stated objectives of increasing average revenue per order while not negatively impacting conversion rate and holding average last mile traversed flat. Note that, the baseline is an LTR model learned on data that is not specially sampled (unlike Sample-Rank).

\subsection{Offline Experiment Results}
AUC \& NDCG: We observe that the amount of change in goal distribution parameters can be correlated with change in AUC and NDCG. Specifically, in Table \ref{tab3} and \ref{tab5} we can see that greater the deviation from baseline parameters, higher the loss in AUC and NDCG. This is expected since more extreme the postulated goal distribution is, there is a greater chance that the recommendations learned by ranking function are less relevant to users. This trade-off between model metrics and goal distribution parameters needs to be considered in decision for choice of model to use.

top@k: Similar to AUC and NDCG, in Tables \ref{tab4} and \ref{tab6} , we find that more extreme the goal distribution, higher the impact in top@k metrics. We also notice that the size of impact on goal metrics is inversely proportional to position in the recommendation. 

\begin{figure}[htbp]
	\centerline{\includegraphics[scale=0.15]{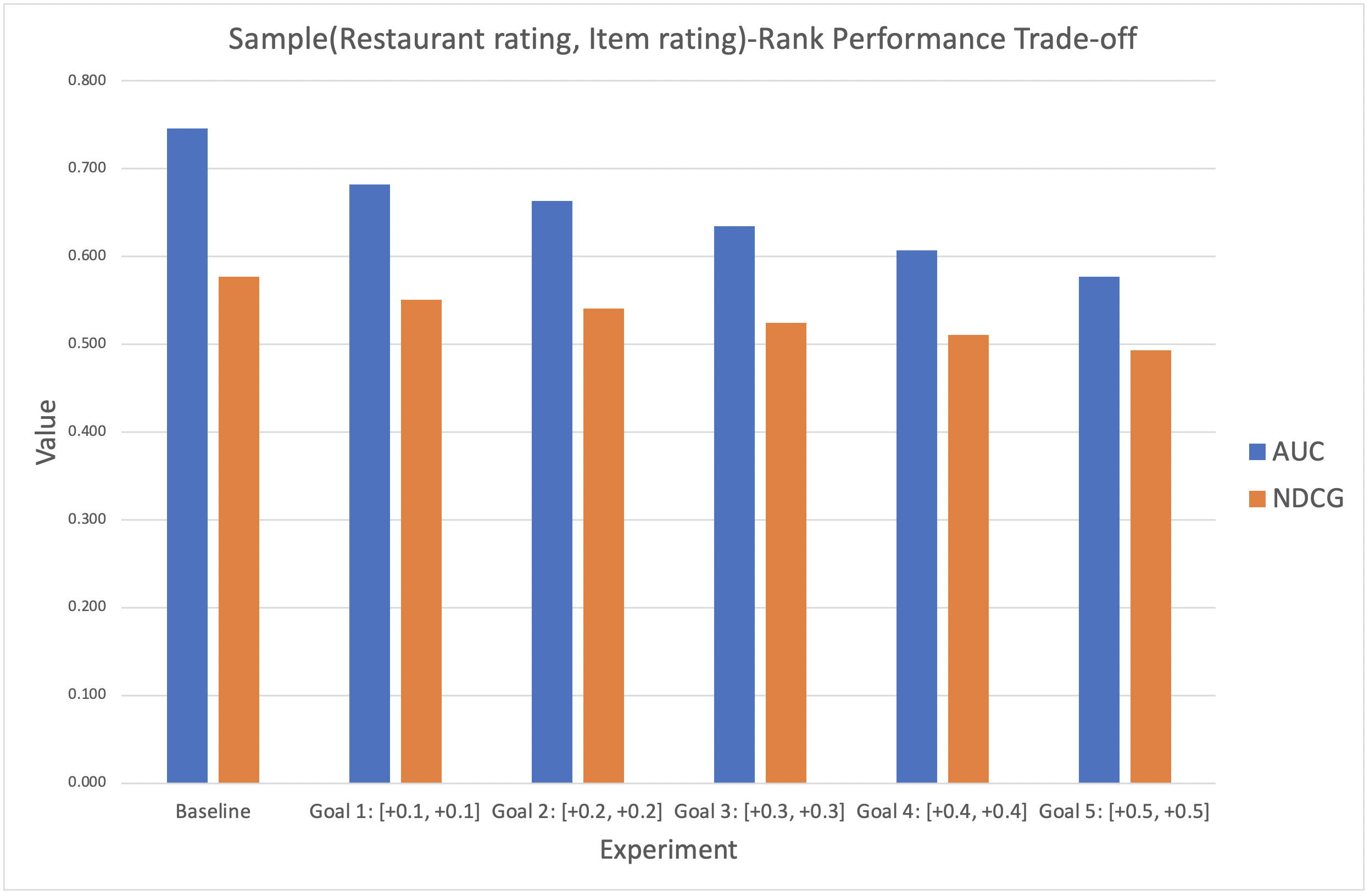}}
	\caption{Offline Ratings Experiment - AUC, NDCG}
	\label{fig1}
\end{figure}

\begin{figure}[htbp]
	\centerline{\includegraphics[scale=0.15]{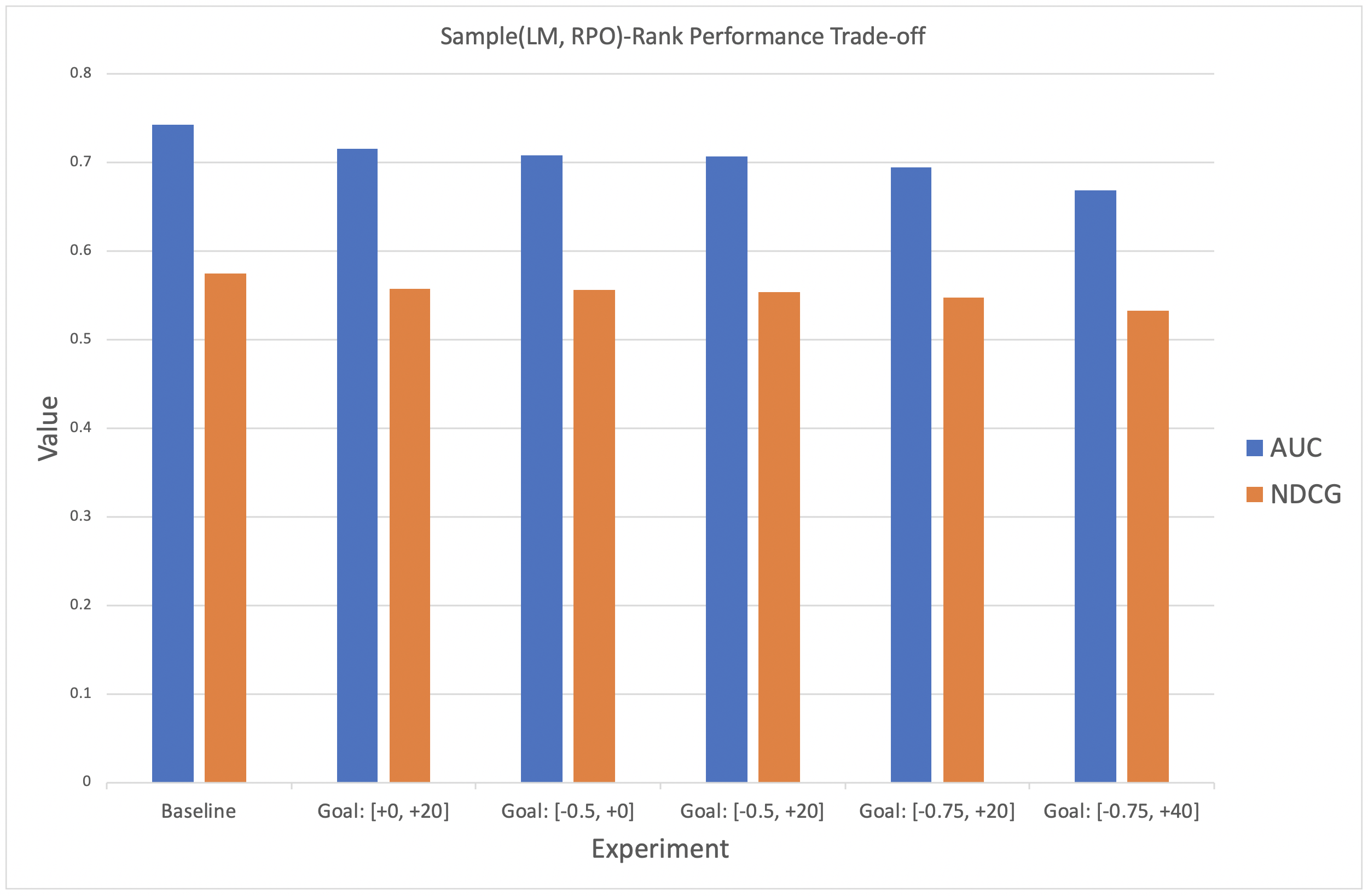}}
	\caption{Offline (LM, RPO) Experiment - AUC, NDCG}
	\label{fig2}
\end{figure}

\begin{table}[htbp]
	\caption{offline ratings experiment - auc, ndcg}
	\begin{center}
	\begin{tabular}{c|c|c}
		\hline
		\textbf{\textit{Experiment}}& \textbf{\textit{AUC}}&\textbf{\textit{NDCG}}\\
		\hline
		Baseline&0.746&0.577\\
		Goal 1&0.682&0.551\\
		Goal 2&0.664&0.541\\
		Goal 3&0.635&0.524\\
		Goal 4&0.607&0.510\\
		Goal 5&0.578&0.494\\
		\hline
	\end{tabular}
		\label{tab3}
	\end{center}
\end{table}

% A table with adjusted row and column spacings
% \setlength sets the horizontal (column) spacing
% \arraystretch sets the vertical (row) spacing
\begingroup
\setlength{\tabcolsep}{5pt} % Default value: 6pt
\renewcommand{\arraystretch}{1.25} % Default value: 1
\begin{table}[htbp]
	\tiny
	\caption{offline ratings experiment - top@k impact}
	\begin{center}
		\begin{tabular}{c|c|c|c|c|c|c|c|c|c|c}
			\hline
			\textbf{\textit{top}}&\multicolumn{2}{|c|}{\textbf{\textit{Goal 1}}}&\multicolumn{2}{|c|}{\textbf{\textit{Goal 2}}}&\multicolumn{2}{|c|}{\textbf{\textit{Goal 3}}}&\multicolumn{2}{|c|}{\textbf{\textit{Goal 4}}}&\multicolumn{2}{|c}{\textbf{\textit{Goal 5}}}\\
			\textbf{\textit{@k}}&\multicolumn{2}{|c|}{\textbf{\textit{[+0.1,+0.1]}}}&\multicolumn{2}{|c|}{\textbf{\textit{[+0.2,+0.2]}}}&\multicolumn{2}{|c|}{\textbf{\textit{[+0.3,+0.3]}}}&\multicolumn{2}{|c|}{\textbf{\textit{[+0.4,+0.4]}}}&\multicolumn{2}{|c}{\textbf{\textit{[+0.5,+0.5]}}}\\
			\hline
			1&0.077&0.094&0.102&0.126&0.124&0.152&0.139&0.164&0.150&0.172\\
			2&0.072&0.094&0.094&0.127&0.113&0.149&0.125&0.158&0.134&0.163\\
			3&0.067&0.090&0.087&0.121&0.102&0.140&0.113&0.148&0.121&0.151\\
			4&0.062&0.086&0.080&0.114&0.092&0.132&0.102&0.138&0.108&0.140\\
			8&0.042&0.064&0.054&0.084&0.061&0.095&0.065&0.094&0.069&0.097\\
			12&0.026&0.044&0.034&0.058&0.038&0.063&0.040&0.061&0.043&0.064\\
			16&0.016&0.029&0.020&0.037&0.023&0.039&0.025&0.038&0.026&0.040\\
			20&0.010&0.019&0.013&0.024&0.014&0.025&0.016&0.024&0.017&0.026\\
			\cline{2-7} 
			\hline
		\end{tabular}
		\label{tab4}
	\end{center}
\end{table}
\endgroup

\begin{table}[htbp]
	\caption{offline (lm, rpo) experiment - auc, ndcg}
	\begin{center}
		\begin{tabular}{c|c|c}
			\hline
			\textbf{\textit{Experiment}}& \textbf{\textit{AUC}}&\textbf{\textit{NDCG}}\\
			\hline
			Baseline&0.743&0.575\\
			Goal 1&0.715&0.557\\
			Goal 2&0.708&0.556\\
			Goal 3&0.707&0.554\\
			Goal 4&0.694&0.548\\
			Goal 5&0.669&0.533\\
			\hline
		\end{tabular}
			\label{tab5}
	\end{center}
\end{table}

% A table with adjusted row and column spacings
% \setlength sets the horizontal (column) spacing
% \arraystretch sets the vertical (row) spacing
\begingroup
\setlength{\tabcolsep}{5pt} % Default value: 6pt
\renewcommand{\arraystretch}{1.25} % Default value: 1
\begin{table}[htbp]
	\tiny
	\caption{offline (lm, rpo) experiment - top@k impact}
	\begin{center}
		\begin{tabular}{c|c|c|c|c|c|c|c|c|c|c}
			\hline
			\textbf{\textit{top}}&\multicolumn{2}{|c|}{\textbf{\textit{Goal 1}}}&\multicolumn{2}{|c|}{\textbf{\textit{Goal 2}}}&\multicolumn{2}{|c|}{\textbf{\textit{Goal 3}}}&\multicolumn{2}{|c|}{\textbf{\textit{Goal 4}}}&\multicolumn{2}{|c}{\textbf{\textit{Goal 5}}}\\
			\textbf{\textit{@k}}&\multicolumn{2}{|c|}{\textbf{\textit{ [-0,+20]}}}&\multicolumn{2}{|c|}{\textbf{\textit{[-0.5,+0]}}}&\multicolumn{2}{|c|}{\textbf{\textit{[-0.5,+20]}}}&\multicolumn{2}{|c|}{\textbf{\textit{[-0.75, +20]}}}&\multicolumn{2}{|c}{\textbf{\textit{[-0.75, +40]}}}\\
			\hline
			1&0.002&5.546&-0.147&-1.518&-0.156&5.811&-0.232&6.179&-0.246&13.167\\
			2&0.003&5.045&-0.141&-1.641&-0.147&5.247&-0.220&5.466&-0.227&12.575\\
			3&0.005&4.378&-0.131&-1.783&-0.137&4.595&-0.203&4.761&-0.207&11.692\\
			4&0.006&3.687&-0.119&-1.866&-0.126&3.966&-0.184&4.048&-0.184&10.644\\
			8&0.001&1.672&-0.078&-1.776&-0.086&1.963&-0.120&2.063&-0.121&6.847\\
			12&-0.004&0.601&-0.050&-1.207&-0.056&0.843&-0.076&1.020&-0.077&4.162\\
			16&-0.006&0.147&-0.031&-0.729&-0.035&0.301&-0.046&0.519&-0.049&2.423\\
			20&-0.007&-0.002&-0.022&-0.392&-0.024&0.070&-0.031&0.292&-0.034&1.418\\
			\cline{2-7} 
			\hline
		\end{tabular}
		\label{tab6}
	\end{center}
\end{table}
\endgroup

\subsection{Online Experiment Results}
\begin{figure}[htbp]
	\centerline{\includegraphics[scale=0.15]{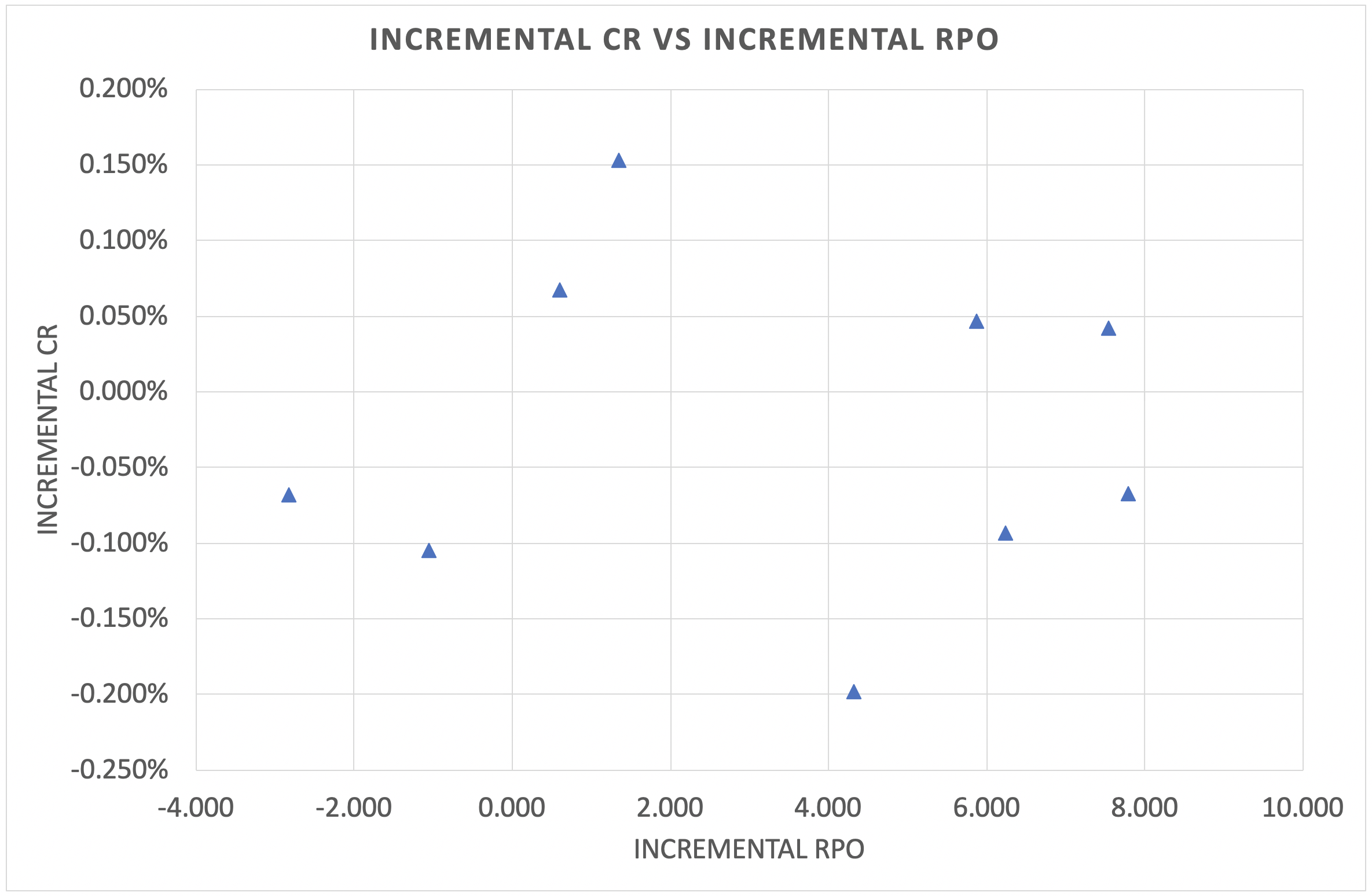}}
	\caption{Online (LM, RPO) Experiment - Incremental CR vs Incremental RPO}
	\label{fig3}
\end{figure}

\begin{figure}[htbp]
	\centerline{\includegraphics[scale=0.15]{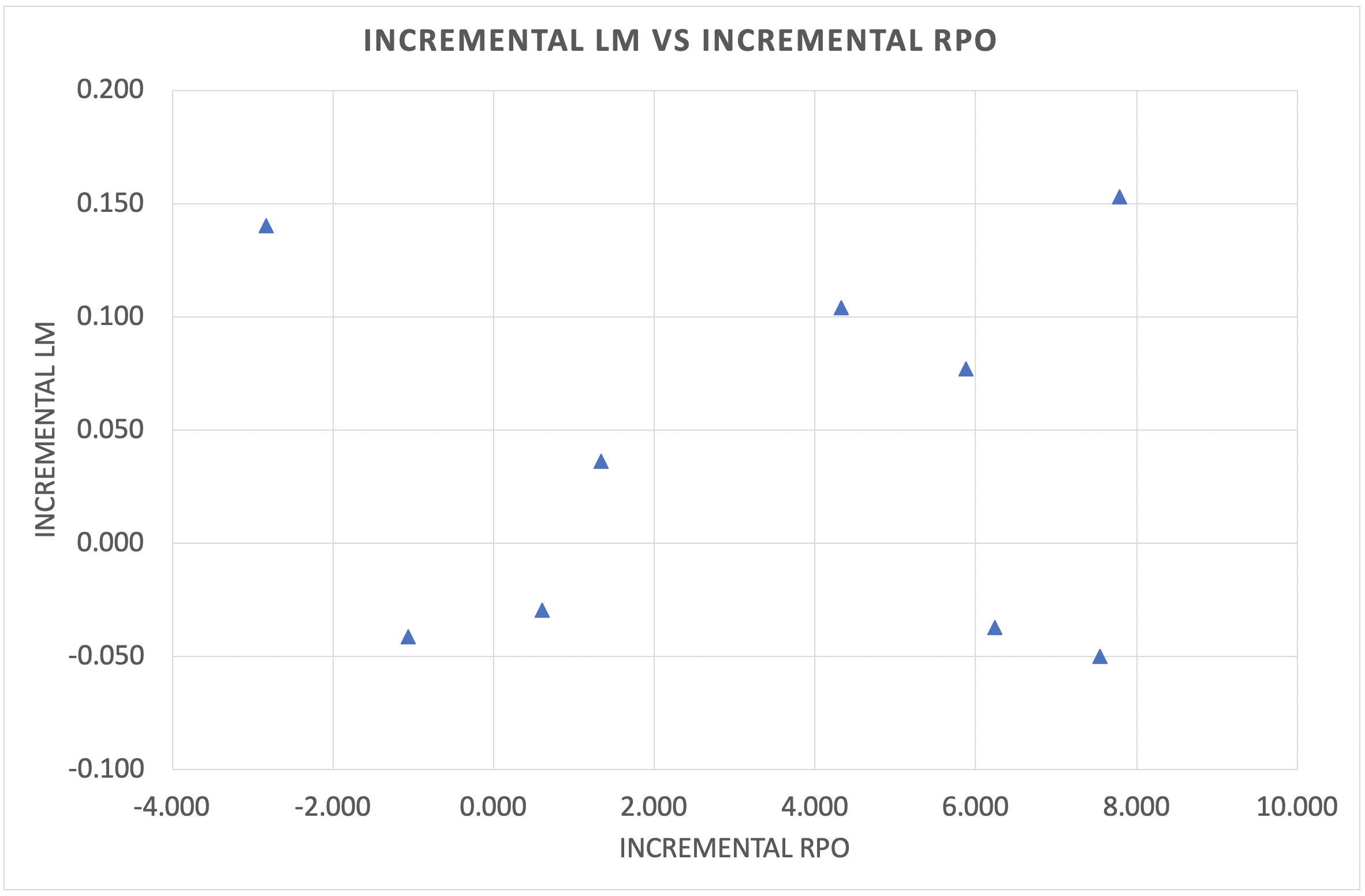}}
	\caption{Online (LM, RPO) Experiment - Incremental LM vs Incremental RPO}
	\label{fig4}
\end{figure}

For the online experiment we deployed the Sample-Rank (LM, RPO) model described in Goal 1 in Table \ref{tab2}. A 50\%-50\% A/B experiment for baseline vs. Sample-Rank was conducted for 9 days in a large market. We observed that the Sample-Rank variant achieved a significant 2.64\% increase (p-value $<$ 0.05) in average revenue per order while having no detrimental impact on the conversion rate and holding LM flat. Table \ref{tab7} shows the daily trend in conversion rate, average revenue per order and LM where the Sample-Rank outperforms the baseline, as can also be observed in Fig. \ref{fig3} and \ref{fig4}.

% A table with adjusted row and column spacings
% \setlength sets the horizontal (column) spacing
% \arraystretch sets the vertical (row) spacing
\begingroup
\setlength{\tabcolsep}{8pt} % Default value: 6pt
\renewcommand{\arraystretch}{1.25} % Default value: 1
\begin{table}[htbp]
	\tiny
	\caption{online (lm, rpo )experiment - incremental LM, RPO \& CR}
	\begin{center}
		\begin{tabular}{c|c|c|c}
			\hline
			%\cline{2-4} 
			\textbf{\textit{Day}}&\textbf{\textit{Incremental LM}}&\textbf{\textit{Incremental RPO}}&\textbf{\textit{Incremental CR(\%)}}\\
			\hline
			1&0.104&4.323&-0.198\%\\
			2&-0.030&0.601&0.067\%\\
			3&0.140&-2.823&-0.068\%\\
			4&-0.041&-1.059&-0.105\%\\
			5&0.077&5.875&0.047\%\\
			6&-0.037&6.244&-0.093\%\\
			7&0.036&1.344&0.153\%\\
			8&0.153&7.789&-0.067\%\\
			9&-0.050&7.547&0.043\%\\
			\hline
		\end{tabular}
			\label{tab7}
	\end{center}
\end{table}
\endgroup

\section{Discussion, Conclusion and Future Work}
In this paper, we introduced a sampling and ranking method, Sample-Rank, for multi-stakeholder recommendation systems. This system does not require explicit formulation of the problem as an MOO provided the stakeholder objectives can be represented as features which are used as parameters to obtain a goal distribution for sampling. We showed promising results of application of this method in a large-scale online experiment in a large online marketplace. This system can be trivially extended to other popular ranking methods and can scale to real-world recommender scales. 
We are pursuing two tracks for future work. First is exploring extending Sample-Rank to data which may not be modeled as Gaussian mixtures. Second is about getting a better understanding of and hence being able to control the impact goal parameters’ configuration has on recommendations.

\end{document}